\documentclass[english]{article}
\usepackage[T1]{fontenc}
\usepackage[utf8]{inputenc}
\usepackage{array}
\usepackage{float}
\usepackage{multirow}
\usepackage{graphicx}
\usepackage[authoryear]{natbib}

\makeatletter

\providecommand{\tabularnewline}{\\}

\usepackage[final]{nips_2017}

\usepackage{hyperref}       
\usepackage{url}            
\usepackage{booktabs}       
\usepackage{amsfonts}       
\usepackage{nicefrac}       
\usepackage{microtype}      

\makeatother

\usepackage{babel}
\begin{document}

\title{Poverty Mapping Using Convolutional Neural Networks Trained on High
and Medium Resolution Satellite Images, With an Application in Mexico}

\author{%
\begin{tabular}{ccc}
Boris Babenko & Jonathan Hersh & David Newhouse\tabularnewline
 Orbital Insight &  Chapman University &  World Bank\tabularnewline
Mountain View, CA & Orange, CA & Washington, DC\tabularnewline
\texttt{boris@orbitalinsight.com} &  \texttt{hersh@chapman.edu}  & \texttt{dnewhouse@worldbank.org} \tabularnewline
 &  & \tabularnewline
Anusha Ramakrishnan & Tom Swartz & \tabularnewline
 World Bank & Orbital Insight & \tabularnewline
Washington, DC & Mountain View, CA & \tabularnewline
 \texttt{aramarishnan2@worldbank.org} &  \texttt{tom@orbitalinsight.com}  & \tabularnewline
\end{tabular}}
\maketitle
\begin{abstract}
Mapping the spatial distribution of poverty in developing countries
remains an important and costly challenge. These ``poverty maps''
are key inputs for poverty targeting, public goods provision, political
accountability, and impact evaluation, that are all the more important
given the geographic dispersion of the remaining bottom billion severely
poor individuals. In this paper we train Convolutional Neural Networks
(CNNs) to estimate poverty directly from high and medium resolution
satellite images. We use both Planet and Digital Globe imagery with
spatial resolutions of 3-5 m$^{2}$ and 50 cm$^{2}$ respectively,
covering all 2 million km$^{2}$ of Mexico. Benchmark poverty estimates
come from the 2014 MCS-ENIGH combined with the 2015 Intercensus and
are used to estimate poverty rates for 2,456 Mexican municipalities.
CNNs are trained using the 896 municipalities in the 2014 MCS-ENIGH.
We experiment with several architectures (GoogleNet, VGG) and use
GoogleNet as a final architecture where weights are fine-tuned from
ImageNet. We find that 1) the best models, which incorporate satellite-estimated
land use as a predictor, explain approximately 57\% of the variation
in poverty in a validation sample of 10 percent of MCS-ENIGH municipalities;
2) Across all MCS-ENIGH municipalities explanatory power reduces to
44\% in a CNN prediction and landcover model; 3) Predicted poverty
from the CNN predictions alone explains 47\% of the variation in poverty
in the validation sample, and 37\% over all MCS-ENIGH municipalities;
4) In urban areas we see slight improvements from using Digital Globe
versus Planet imagery, which explain 61\% and 54\% of poverty variation
respectively. We conclude that CNNs can be trained end-to-end on satellite
imagery to estimate poverty, although there is much work to be done
to understand how the training process influences out of sample validation. 

{\let\thefootnote\relax\footnotetext{Presented at NIPS 2017 Workshop on Machine Learning for the Developing World}} 
\end{abstract}

\section{Introduction}

Understanding the spatial distribution of poverty is an important
step before poverty can be eradicated worldwide. In recent decades,
we have seen dramatic declines of poverty in many areas including
India, China, and many areas of East Asia, Latin America and Africa.
There is much to celebrate in this decline, as billions of people
have risen out of poverty. The poverty that remains – the roughly
1 billion individuals worldwide below the international poverty line
of \$1.90 per day – are distributed non-uniformly across space, often
in rural and urban ``pockets'' that are inaccessible and frequently
changing. It's posited that these areas are unlikely to integrate
with the path of the global economy unless policy measures are taken
to ensure their integration. 

The first step to addressing this poverty is knowing with precision
where it is located. Unfortunately, this has proven to be a non-trivial
task. The standard method of generating a geographic distribution
of poverty – a ``poverty map'' – involves combining a household
consumption survey with a broader survey, typically a Census. While
this method is accurate enough to produce official statistics (Elbers,
et al., 2003), it has several disadvantages. Censuses and consumption
surveys are expensive, costing millions for countries to produce.
The lag time between survey and production of poverty rates can be
several years due to the time needed to collect, administer, and produce
statistics on poverty rates. Finally, because of security concerns
and geographic remoteness, it is often infeasible to survey every
area within a country. 

The combination of computer vision trained against satellite imagery
holds much promise for the creation of frequently updated and accurate
poverty maps. Several research groups have explored the capabilities
of computer vision trained against satellite imagery to estimate poverty.
Jean et al. (2015) use a transfer learning approach that uses the
penultimate layer of a CNN trained against night time lights as explanatory
variables to estimate poverty. Engstrom et al. (2017) use intermediate
features (cars, roofs, crops) identified through computer vision to
estimate poverty. This paper takes the direct route and estimates
an end-to-end CNN trained to estimate poverty rates of urban and rural
municipalities in Mexico. We complement these by incorporating land
use estimates estimated from Planet imagery. The results are modest
but encouraging. The best models, which incorporate land use as a
predictor, explain 57\% of poverty in a 10\% validation sample. However,
looking at all MCS-ENIGH municipalities, the explanatory power drops
to 44\%. We speculate as to why we see this decline out of the validation
sample and suggest some possible improvements. 

\section{Data}

\subsection{Mexican Survey Data}

The CNN is trained using survey data from the 2014 MCS-ENIGH. Poverty
benchmark data is created using a combination of the 2014 MCS-ENIGH
household survey, the second from the 2015 Intercensus. The 2014 MCS-ENIGH
survey covers 58,125 households, of which approximately 75\% are urban
and 25\% are rural. The survey samples 896 municipalities out of roughly
2,500. The survey collects income per adult equivalent, which is the
income metric used to calculate the official poverty rate. The 2015
Intercensus is a survey of households conducted every 5 years. For
2015 the Intercensus samples 229,622 households. The Intercensus contains
only household labor income and transfer income, and not total household
income. However, labor income and household income are strongly linearly
correlated, with an $R^{2}$ value of approximately 0.9. We experimented
with different samples from the Intercensus to determine whether number
of household data points on which the CNN is trained affects performance
accuracy. 

We considered two separate poverty rates: the minimum well-being poverty
line and the well-being poverty line. These poverty lines varied for
urban and rural areas. For each administrative unit we calculated
the fraction of households living in poverty. Thus the end-to-end
prediction task beings with satellite imagery and ends with a prediction
for each administrative area of the distribution over three ``buckets'':
below minimum well-being, between minimum well-being and well-being,
and above well-being. 

\subsection{Satellite Imagery}

We used satellite imagery provided by both Planet and Digital Globe,
examples of which is shown in figure 1. Assessing the comparative
tradeoffs between Planet and Digital was one of the goals of the project.
Digital Globe imagery is of higher resolution, with spatial resolution
of 50 cm$^{2}$, and covers the years 2014-2015. Planet imagery varies
in resolution between 3 - 5 m$^{2}$ and ranged in date from late
2015 to early 2017. Digital Globe imagery is only used in urban areas,
as coverage in rural areas is sparse. Planet imagery is ``4-band'',
and includes the near-infrared (NIR) band, while the Digital Globe
imagery does not. We experimented with including the NIR band during
the training process, but ultimately saw better results with the exclusion
of this band. 

\begin{figure}[H]
\centering{} \includegraphics[scale=0.4]{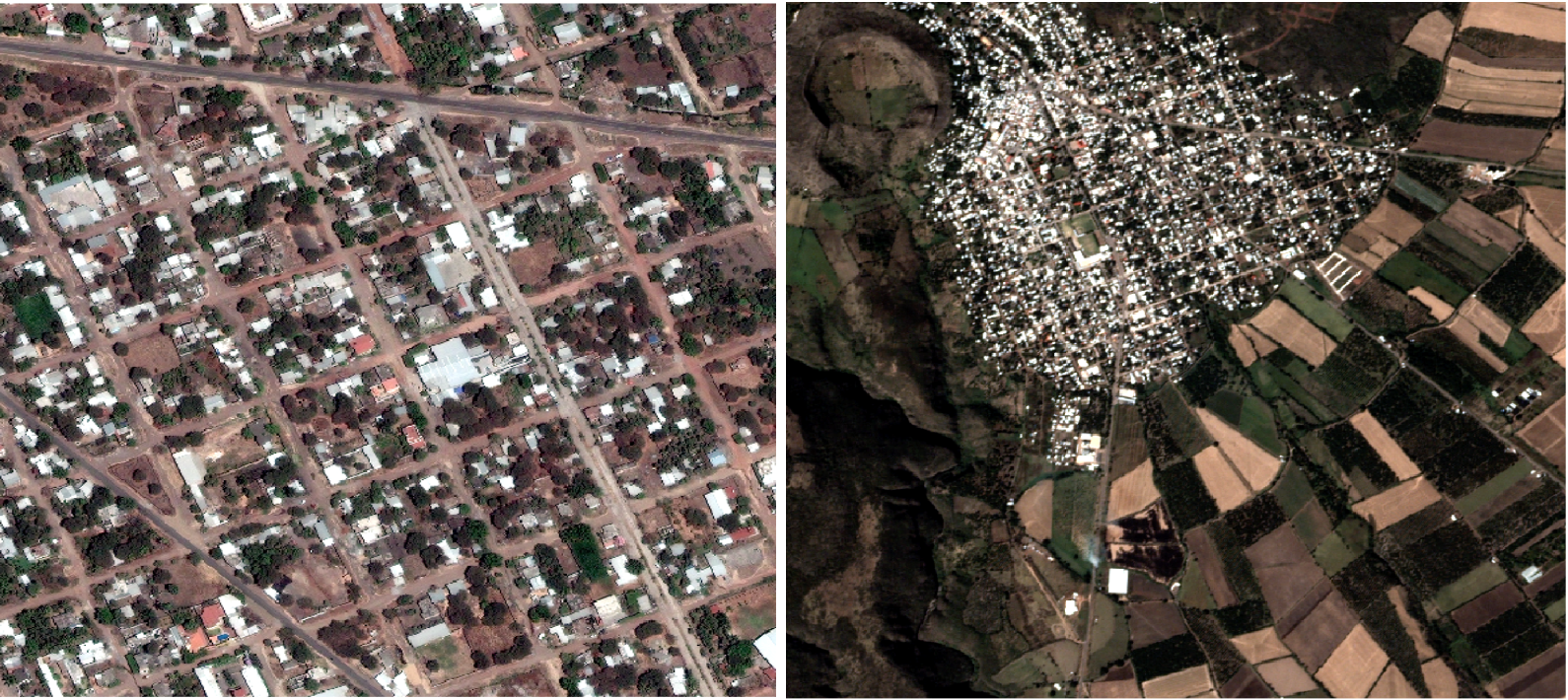}\label{planet_DG}\caption{Example Digital Globe (left) and Planet (right) imagery, Michoacan,
(Satellite images (c) 2017 Digital Globe, Inc., Planet)}
\end{figure}

\section{Technical Methodology}

During the training process we experimented with two CNN architectures.
The first is GoogleNet (Szegedy et al., 2014) and the second is a
variant of VGG (Simoyan, 2014). We experimented with various solvers
and weight initializations which were evaluated against an internal
development or ``dev'' set. According to tests using this dev set,
the GoogleNet architecture outperformed the VGG architecture. We also
experimented with fine-tuning the weights of the GoogleNet models.
We compared fine-tuned models, using weights initialized at the values
of a model trained against ImageNet. 

\paragraph{\textmd{Digital Globe and Planet imagery both include three bands
of Red, Blue and Green (RGB) values. Planet imagery includes a 4th
band for near infrared. We experimented with training models to include
this additional information. The ImageNet dataset consists only of
RGB imagery, so it is not-trivial to fine-tune from an ImageNet-trained
model to a model with a 4-band input. Therefore, for 4-band input
we trained from scratch and only attempted fine-tuning from ImageNet
for 3-band versions of the imagery. That is to say, ultimately the
NIR band was dropped. }}

\section{Results}

Focusing on the urban subsample, table \ref{Urban_Enigh_Trained-1}
presents the CNN predictions for urban areas using imagery for either
Digital Globe or Planet, using the 10\% withheld validation sample.
We present $R^{2}$ estimates that show the correlation between predicted
poverty and benchmark poverty as measured in the 2015 Intercensus.
$R^{2}$ is estimated at 0.61 using the Digital Globe imagery, and
0.54 using Planet imagery. Recall we can only compare urban areas
due to lack of coverage of rural areas for Digital Globe. The drop
in performance is modest but not severe, especially considering that
Planet imagery offers daily revisit rates of the earth's landmass.
Poverty estimates for urban areas in Mexico are mapped shown in figure
2. 

Table \ref{Urban_Enigh_Trained} shows the model performance varying
the subsample to include more than the 10\% validation sample. In
the 10\% validation sample, using CNN predictions, we estimate an
$R^{2}$ value between predicted and true poverty between 0.47 and
0.54. When adding landcover classification, estimated via Planet imagery,
to the CNN predictions we estimate an $R^{2}$ value between 0.57
and 0.64. However, when we compare this to estimates within all MCS-ENIGH
areas, the coefficient of variation falls to 0.4 and 0.44 in urban
and rural areas respectively. Outside of the 896 municipalities that
comprise the MCS-ENIGH survey we see explanatory power fall precipitously,
to roughly 0.3. The poor performance outside of MCS-ENIGH municipalities
is puzzling. This could be due to weighting tiles by geographic area
instead of population. It could also be due to MCS-ENIGH municipalities
having differential characteristics from non-MCS-ENIGH municipalities,
such as more homogeneous population density or differential size. 

\begin{table}[t]
\centering{}\caption{Comparison Digital Globe versus Planet Imagery, 10\% Validation Sample}
\label{Urban_Enigh_Trained-1} {\tiny{}}%
\begin{tabular}{>{\raggedright}b{0.08\paperwidth}>{\centering}m{0.17\paperwidth}>{\centering}m{0.15\paperwidth}>{\centering}b{0.07\paperwidth}}
\textbf{\small{}Sample} & \textbf{\small{}$R^{2}$ CNN Predictions using Digital Globe imagery } & \textbf{\small{}$R^{2}$ CNN Predictions using Planet imagery } & \textbf{\small{}\# municipalities }\tabularnewline
\hline 
Urban areas  & 0.61 & 0.54 & 58\tabularnewline
\hline 
\end{tabular}
\end{table}

\begin{figure}[H]
\begin{centering}
\includegraphics[scale=0.4]{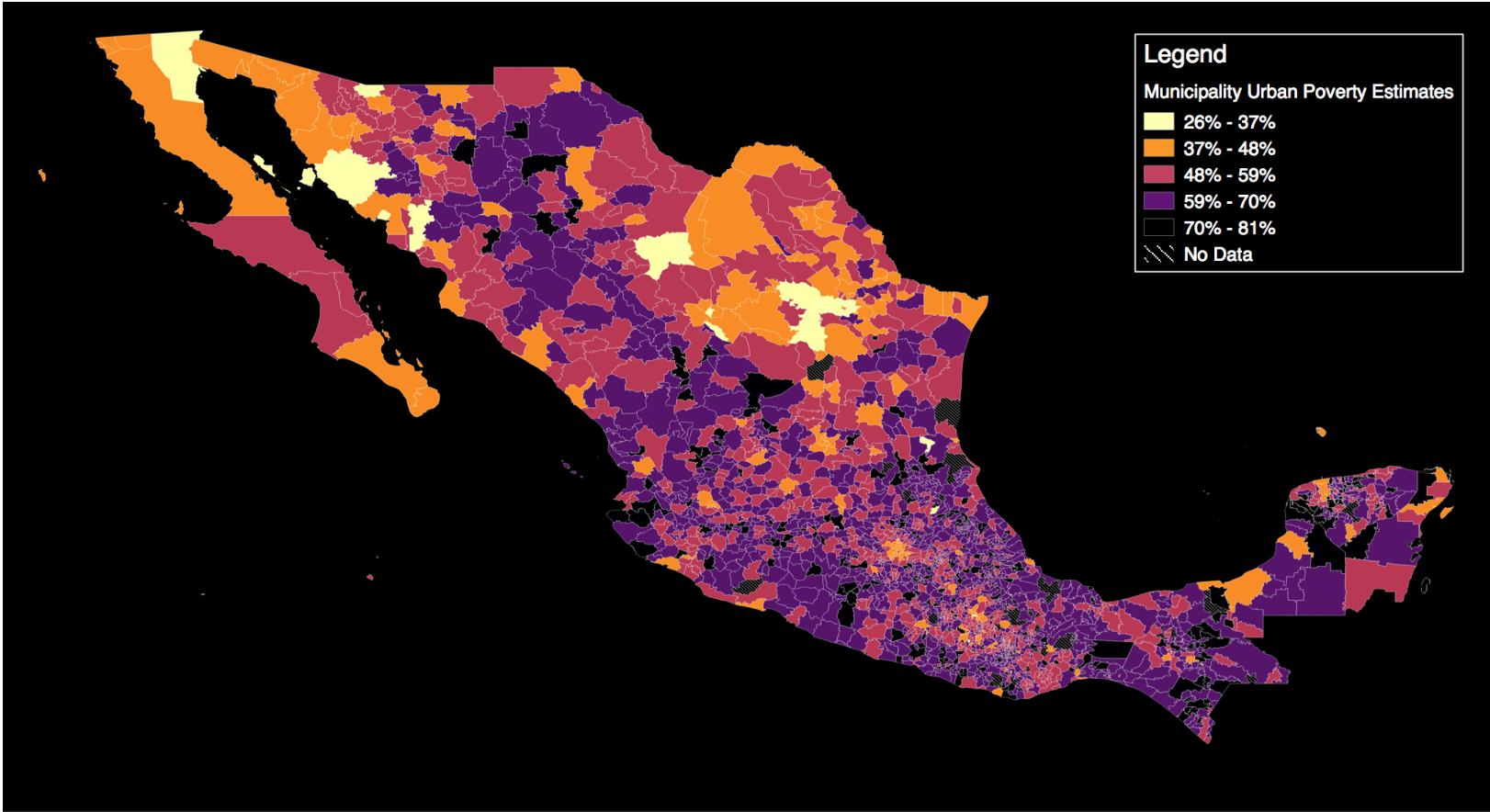}
\par\end{centering}
\centering{}\label{planet_DG-1}\caption{Poverty Estimates, Urban Municipalities}
\end{figure}

\begin{table}[t]
\centering{}\caption{CNN Predictions In and Out of Sample}
\label{Urban_Enigh_Trained} {\tiny{}}%
\begin{tabular}{>{\raggedright}b{0.1\paperwidth}|>{\centering}m{0.05\paperwidth}>{\centering}m{0.08\paperwidth}>{\centering}m{0.08\paperwidth}>{\centering}m{0.08\paperwidth}>{\centering}b{0.07\paperwidth}}
\multicolumn{1}{>{\raggedright}b{0.1\paperwidth}}{\textbf{\small{}Validation}} & \textbf{\small{}Sample} & \textbf{\small{}$R^{2}$ CNN Predictions} & \textbf{\small{}$R^{2}$ Landcover } & \textbf{\small{}$R^{2}$ Both} & \textbf{\small{}Areas }\tabularnewline
\hline 
\multirow{3}{0.1\paperwidth}{{\small{}10\% MCS-ENIGH Validation}} & {\small{}All } & {\small{}0.47} & {\small{}0.49} & {\small{}0.57} & {\small{}109}\tabularnewline
 & {\small{}Urban} & {\small{}0.54} & {\small{}0.52} & {\small{}0.64} & {\small{}58}\tabularnewline
 & {\small{}Rural} & {\small{}0.47} & {\small{}0.49} & {\small{}0.64} & {\small{}51}\tabularnewline
\multicolumn{1}{>{\raggedright}b{0.1\paperwidth}}{} &  &  &  &  & \tabularnewline
\multirow{3}{0.1\paperwidth}{{\small{}all MCS-ENIGH areas}} & {\small{}All } & {\small{}0.37} & {\small{}0.37} & {\small{}0.44} & {\small{}1115}\tabularnewline
 & {\small{}Urban} & {\small{}0.34} & {\small{}0.31} & {\small{}0.4} & {\small{}619}\tabularnewline
 & {\small{}Rural} & {\small{}0.38} & {\small{}0.34} & {\small{}0.44} & {\small{}496}\tabularnewline
\multicolumn{1}{>{\raggedright}b{0.1\paperwidth}}{} &  &  &  &  & \tabularnewline
\multirow{3}{0.1\paperwidth}{{\small{}non MCS-ENIGH areas}} & {\small{}All } & {\small{}0.15} & {\small{}0.23} & {\small{}0.28} & {\small{}2834}\tabularnewline
 & {\small{}Urban} & {\small{}0.06} & {\small{}0.19} & {\small{}0.21} & {\small{}944}\tabularnewline
 & {\small{}Rural} & {\small{}0.22} & {\small{}0.25} & {\small{}0.31} & {\small{}1890}\tabularnewline
\hline 
\end{tabular}
\end{table}

\section*{References}

{\small{}{[}1{]} Elbers, Chris, Jean O. Lanjouw, and Peter Lanjouw.
\textquotedbl{}Micro–level estimation of poverty and inequality.\textquotedbl{}
Econometrica 71, no. 1 (2003): 355-364.}{\small \par}

{\small{}{[}2{]} Engstrom, R., Hersh, J., Newhouse, D. “Poverty from
space: using high resolution satellite imagery for estimating economic
well-being and geographic targeting.” (2016).}{\small \par}

{\small{}{[}3{]} Jean, Neal, Marshall Burke, Michael Xie, W. Matthew
Davis, David B. Lobell, and Stefano Ermon. \textquotedbl{}Combining
satellite imagery and machine learning to predict poverty.\textquotedbl{}
Science 353, no. 6301 (2016): 790-794.}{\small \par}

{\small{}{[}4{]} Szegedy, Christian, Wei Liu, Yangqing Jia, Pierre
Sermanet, Scott Reed, Dragomir Anguelov, Dumitru Erhan, Vincent Vanhoucke,
and Andrew Rabinovich. \textquotedbl{}Going deeper with convolutions.\textquotedbl{}
In Proceedings of the IEEE conference on computer vision and pattern
recognition, pp. 1-9. 2015.}{\small \par}

{\small{}{[}5{]} Simonyan, Karen, and Andrew Zisserman. \textquotedbl{}Very
deep convolutional networks for large-scale image recognition.\textquotedbl{}
arXiv preprint arXiv:1409.1556 (2014).}{\small \par}
\end{document}